\documentclass[letterpaper, 10 pt, conference]{ieeeconf}
\usepackage{cite}
\usepackage{amsmath,amssymb,amsfonts} 
\usepackage{graphicx}
\usepackage{textcomp} 
\usepackage{multirow}
\usepackage{gensymb}
\usepackage{xcolor} 
\usepackage[ruled,vlined]{algorithm2e}
\usepackage{gensymb}
\usepackage{textcomp} 
\usepackage{xcolor} 
\usepackage{color}
\usepackage{bm}
\usepackage{multirow}
\usepackage{graphicx}
\usepackage{subfigure}
\usepackage[mathscr]{eucal}

\IEEEoverridecommandlockouts                              

\makeatletter

\makeatletter
\renewcommand{\maketag@@@}[1]{\hbox{\m@th\normalsize\normalfont#1}}%

\setbox0\hbox{$\xdef\scriptratio{\strip@pt\dimexpr
    \numexpr(\sf@size*65536)/\f@size sp}$}

\newcommand{\scriptveryshortarrow}[1][3pt]{{%
    \vcenter{\hbox{\rule[\scriptratio\dimexpr-.2pt\relax]
               {\scriptratio\dimexpr#1\relax}{\scriptratio\dimexpr.4pt\relax}}}%
   \mkern-4mu\hbox{\let\f@size\sf@size\usefont{U}{lasy}{m}{n}\symbol{41}}}}

\makeatother

\title{\LARGE \bf
An Extrinsic Calibration Method between LiDAR and GNSS/INS for Autonomous Driving
}

\author{Guohang Yan, Jiahao Pi, Chengjie Wang, Xinyu Cai and Yikang Li$^{\dagger}$ 
\thanks{$^{\dagger}$ Corresponding author.}
\thanks{Guohang Yan, Jiahao Pi, Chengjie Wang, Xinyu Cai and Yikang Li are with Autonomous Driving Group, Shanghai AI Laboratory, China. {\tt\small \{yanguohang, pijiahao, wangchengjie, caixinyu,  liyikang\}@pjlab.org.cn}}
}
    
\begin{document}
 
\maketitle

\begin{abstract}
Accurate and reliable sensor calibration is critical for fusing LiDAR and inertial measurements in autonomous driving. This paper proposes a novel three-stage extrinsic calibration method between LiDAR and GNSS/INS for autonomous driving. The first stage can quickly calibrate the extrinsic parameters between the sensors through point cloud surface features so that the extrinsic can be narrowed from a large initial error to a small error range in little time.
The second stage can further calibrate the extrinsic parameters based on LiDAR-mapping space occupancy while removing motion distortion. In the final stage, the z-axis errors caused by the plane motion of the autonomous vehicle are corrected, and an accurate extrinsic parameter is finally obtained. Specifically, This method utilizes the planar features in the environment, making it possible to quickly carry out calibration. Experimental results on real-world data sets demonstrate the reliability and accuracy of our method. The codes are open-sourced on the Github website. The code link is https://github.com/OpenCalib/LiDAR2INS.
\end{abstract}


\section{INTRODUCTION}
Autonomous driving technology has attracted more and more attention with the continuous development of science and technology. Accurate and reliable location information is becoming indispensable to realizing the various complex functions of self-driving cars. Therefore, most self-driving cars will be equipped with the GNSS/INS (Inertial Navigation System) devices, such as NovAtel, Trimble, and other high-precision positioning devices. LiDAR (Light Detection and Ranging) is another crucial sensor in autonomous driving technology. SLAM (Simultaneous Localization and Mapping) and object detection are the two most important applications of LiDAR in the field of autonomous driving. The main role of SLAM is mapping and localization. It should be noted that each laser point of LiDAR is generated at a different reference pose during the motion of the self-driving car, which is also the source of LiDAR motion distortion. In autonomous driving, the motion error of the laser frame caused by motion cannot be ignored. Usually, the motion distortion of LiDAR is removed with the help of GNSS/INS. High-precision map construction and localization highly depend on LiDAR and GNSS/INS fusion. Accurate 3D-LiDAR and GNSS/INS extrinsic calibration are essential to determine their coordinate relationship and perform sensor fusion. 

\begin{figure}[ht]
\centering
\includegraphics[scale=0.4]{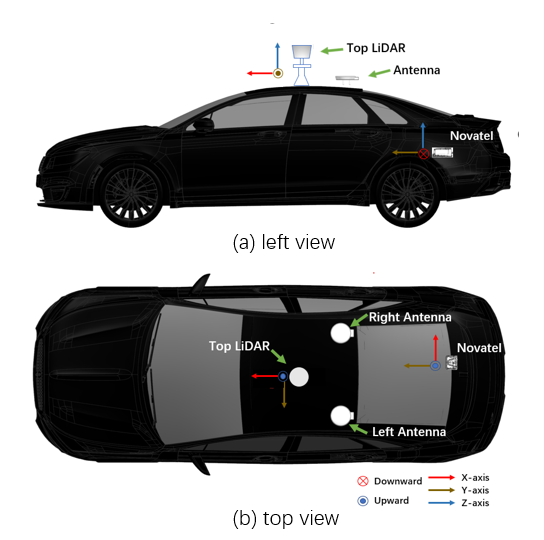}
\caption{(a) Left view of our experimental platform. (b) Top view of our experimental platform.}
\label{Fig.setup}
\end{figure}

There are few methods for specifically calibrating LiDAR to GNSS/INS and even fewer open-source methods. Lidar-align \cite{lidaralign2019} is a simple open-source method for finding the extrinsic calibration between a 3D LiDAR and a 6-dof pose sensor, which also pointed out that accurate results require highly non-planar motions. This makes the technique poorly suited for calibrating sensors mounted to cars. Currently, most of the LiDAR to GNSS/INS calibration methods still refer to the calibration method of LiDAR-IMU. Due to the rapid development of LiDAR-IMU systems in recent years \cite{zhang2014loam, shan2018lego, zuo2020lic, ye2019tightly}, many methods have also emerged to calibrate LiDAR-IMU \cite{le20183d, lyu2019efficient, lvtargetless, mishra2021target, li20213d, lv2022observability}. These methods generally calibrate the intrinsic parameters of the IMU and the extrinsic parameters of the LiDAR-IMU for small robot platforms. However, these calibration methods do not necessarily work well when directly used for calibrating LiDAR to GNSS/INS in autonomous vehicles. 
Currently, there is no calibration method from LiDAR to GNSS/INS specifically for autonomous driving. Therefore, we propose a calibration method from LiDAR to GNSS/INS specifically for autonomous driving applications in this paper. Inspired by \cite{liu2021balm}, we propose a rough calibration method based on the constraints imposed on the corresponding surface feature points between multiple frames. Subsequently, an optimization equation based on LiDAR-mapping space occupancy is used to further refine the rough calibration results. At the same time, the errors that cannot be evaluated in the z-axis caused by the plane motion of the vehicle are corrected through multiple fiducial points. Through this three-stage method, an accurate extrinsic parameter that overcomes the dimension of plane motion is gradually obtained. 
Our method is less demanding in the environment setting (a calibration scene only needs to measure the fiducial points once) and is a fully automatic calibration procedure, while the time required for calibration is relatively very small, so the proposed method can contribute to a more efficient and practical large-scale deployment.

The contributions of this work are listed as follows: 
\begin{enumerate}
\item We introduce an algorithm for solving LiDAR and GNSS/INS extrinsic calibration initialization by directly minimizing the distance from the feature point to the plane. 
\item An octree-based space occupancy refinement method is defined to further refine the extrinsic parameters for improving LiDAR's mapping quality through the pose-sensor. The calibration accuracy of the z-axis is improved through fiducial points matching.
\item 
Evaluated on real-world datasets, we quantitatively and qualitatively demonstrate our method's robustness and accuracy; meanwhile, the related codes have been open-sourced on GitHub.  
\end{enumerate}

\section{RELATED WORK}
Researchers have proposed many methods to solve the multi-sensor calibration problem. Sensor calibration can be divided into two parts: intrinsic parameter calibration and extrinsic parameter calibration, and intrinsic parameters determine the internal mapping relationship of the sensor. For example, the IMU intrinsic parameters include the zero bias of gyroscope and accelerometer, scale factor, and installation error, which can be calibrated by method \cite{li2014high, rehder2016extending, yang2020online, liu2019error, liu2019novel}. The extrinsic parameters determine the transformation relationship between the sensor and the external coordinate system, including 6 degrees of freedom parameters of rotation and translation. Lidar-align \cite{lidaralign2019} proposed a extrinsic calibration method that minimizes the distance between each point of the stitching point cloud by GNSS/INS and its nearest neighbour. However, this method is runtime and cannot solve the z-axis error caused by plane motion. Currently, most LiDAR to GNSS/INS calibration methods still refer to the calibration method of LiDAR-IMU. LiDAR-IMU extrinsic is usually calibrated together with IMU intrinsic calibration, e.g. methods \cite{le20183d, lyu2019efficient, lvtargetless, mishra2021target, li20213d, lv2022observability}. Usually, LiDAR to GNSS/INS calibration removes the intrinsic's calibration part and only calibrates the extrinsic.

LV et al. \cite{lvtargetless} proposed the first open-source LIDAR-IMU calibration toolbox based on continuous-time batch estimation. This method can correct the offset of the coordinate system in different positions, proving the method's reliability. Subsequently, they extended this method and proposed OA-LICalib \cite{lv2022observability}, which seeks to automatically select the most informative data segment for calibration, avoiding some data without sufficient motion or scene constraints, thus improving calibration accuracy and reducing the computational cost. Gentil et al. \cite{li20213d} used gauss equation regression to eliminate IMU motion distortion, and the optimization method based on the factor graph is used to calibrate LIDAR and IMU. Jiao et al. \cite{jiao2019automatic} obtained the initial value of calibration through hand-eye calibration \cite{horaud1995hand}. Then, the appearance-based method is used to optimize the obtained parameters by minimizing the residual function composed of feature points to the plane. Mishra et al. \cite{mishra2021target} proposed an optimization scheme based on EKF to calibrate LIDAR and IMU.
\cite{furgale2013unified, rehder2014spatio} proposed a spatio-temporal calibration method using the continuous-time batch estimation framework for the camera-IMU calibration.
Forster et al. \cite{forster2015manifold} solved a non-linear batch estimation problem to determine the unknown extrinsic calibration parameter. The above calibration methods need to be based on good initial values. Park et al. \cite{park2020spatiotemporal} applied the calibration from coarse to fine, first estimating the closed-form solution and then batch optimizing the continuous-time trajectory to obtain more accurate results.

Due to the lack of Z-direction motion, the existing calibration methods are usually difficult to correct the offset of the z-axis. At the same time, self-driving cars usually require a fast and precise calibration.
To this end, we propose a calibration method from LiDAR to GNSS/INS specifically for autonomous driving applications. Our work represents a coarse to fine calibration for LIDAR and GNSS/INS. After getting a satisfactory result, our method also corrects the deviation of the z-axis through fiducial points matching. And our whole calibration procedure takes very little time.

\section{METHODOLOGY}
\begin{figure*}[ht]
\centering
\includegraphics[scale=0.72]{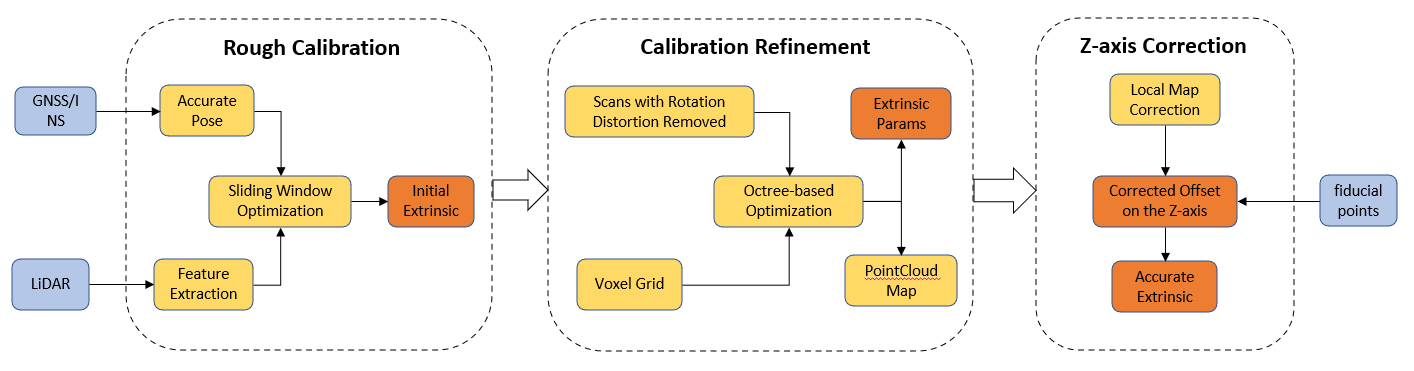}
\caption{The pipeline of the proposed LIDAR and pose-sensor extrinsic calibration method includes three parts: rough calibration, calibration refinement and Z-axis correction.}
\label{Fig.pipeline}
\end{figure*}
This section introduces the details of our method, including rough calibration, calibration refinement, and z-axis correction. Fig.~\ref{Fig.pipeline} shows the overview of the proposed method. 
\begin{figure}[ht]
\centering
\includegraphics[scale=0.17]{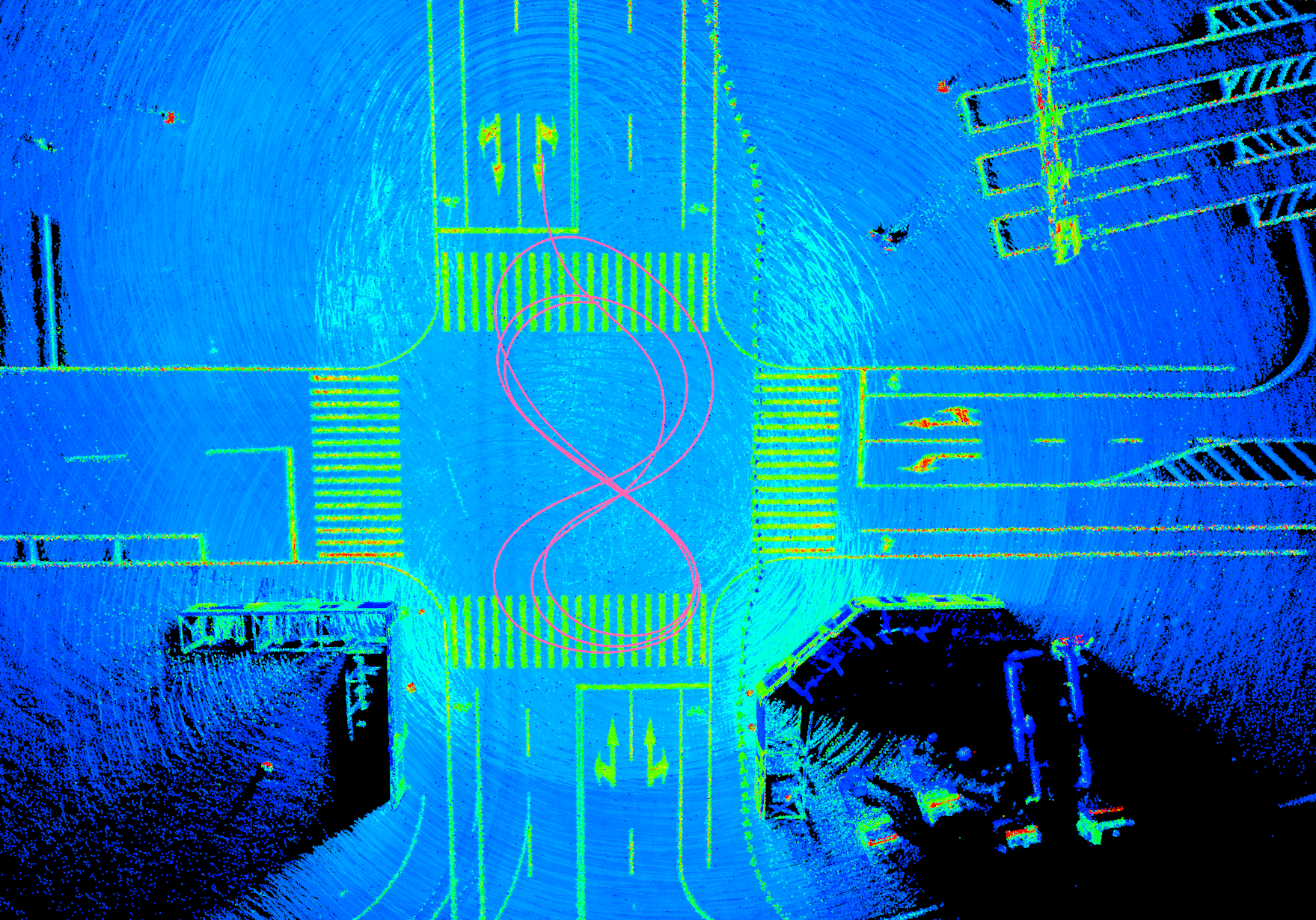}
\caption{The calibration data collection method is that the self-driving car circles three “8” characters around the intersection, and the trajectory in the figure is the driving trajectory when the vehicle collects data.}
\label{Fig.data_collect}
\end{figure}

\subsection{Problem Formulation}
As shown in Fig.~\ref{Fig.data_collect}, the vehicle collects LiDAR and GSNN/INS sequence data at the intersection by walking three figure-8-shape trajectories and keeps the vehicle speed between 10 km/h and 20 km/h. Then, the GNSS/INS pose data corresponding to the LiDAR timestamp is obtained through the data processing module. With accurate extrinsic parameters, we can reconstruct the surrounding environment based on LiDAR and GNSS/INS data. Therefore, Our goal is to find a rigid transformation $\bm{T}=\{\bm{R,t}\}$ from LiDAR to GNSS/INS so that the 3D reconstruction results obtained by splicing the multi-frame point clouds through the pose provided by GNSS/INS are more accurate. $\bm{R}$ is a 3D rotation, $\bm{R} \in SO(3)$ and $\bm{t}$ is a 3D translation, $\bm{t} \in R^{3}$. First, we process the LiDAR point cloud pose $T^L$ in the LiDAR coordinate system and convert it to obtain the pose $T^I$ of the point cloud in the GNSS/INS coordinate system by the extrinsic parameters $\bm{T_{I}^L}$. Then we get the pose of point cloud in the world coordinate system through GNSS/INS output pose $T_{W}^I$
\begin{equation}\label{formulated problem}
\begin{aligned}
T_{i}^W= (T_{W}^I)_i\bm{T_{I}^L} T_{i}^L = \{R_{i}^W,t_{i}^W\}
\end{aligned} 
\end{equation}
Then, recover the 3D reconstruction map $\bm{M}$ through pose  $T^W$ and LiDAR point cloud sequence $\bm{P^L}$.
\begin{equation}\label{formulated problem}
\begin{aligned}
\bm{M} = \sum\limits _{i=1}^{N} (R_{i}^W \bm{P_{i}^L} + t_{i}^W)
\end{aligned} 
\end{equation}
Where $\bm{M}$ is the global 3D point-cloud map in the world coordinate system, we aim to find an extrinsic parameter $\bm{T_{I}^L}$ to make the reconstructed map $\bm{M}$ has the best quality.


\subsection{Rough Calibration}
The first step is rough calibration. Rough calibration aims to quickly reduce the extrinsic parameter from an initial value with a large error to a small error range. 
In our experiments, our rough calibration can reduce the extrinsic error in angle and translation from over 20° and 0.5m compared to the ground truth to within 0.2° and 0.03m in less than 20s. In order to reduce the runtime, we perform feature extraction on the point cloud. Similar to method \cite{liu2021balm}, we extract the plane features in the point cloud features through adaptive voxelization. The next step is to project the point cloud into the same coordinate system combined with the pose information of GNSS/INS and then optimize it in a sliding window. The specific method of optimization is to assume that there is a sliding window of $n$ frames, in which the pose of GNSS/INS is denoted as $T_{1}^W,..,T_{n}^W$. 
In the sliding window, we first extract the point cloud's feature through each point cloud's curvature. 
\begin{figure}[ht]
\centering
\includegraphics[scale=0.3]{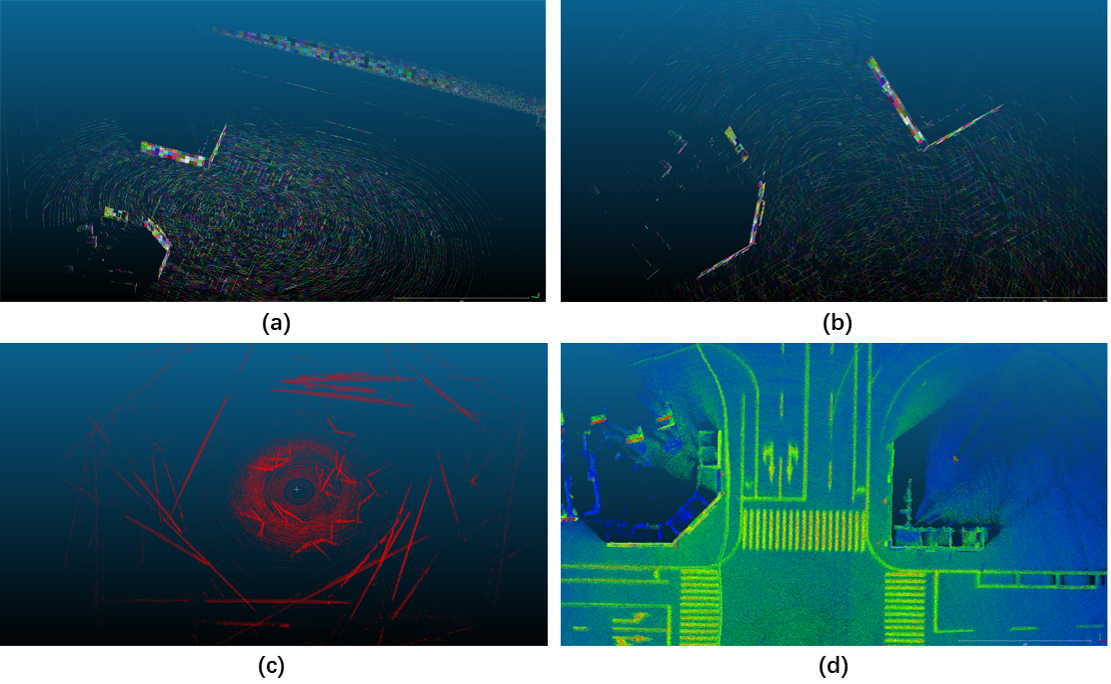}
\caption{(a) and (b) are the feature maps of the extracted point cloud surface. (c) The feature map point cloud is extracted in the initial state. (d) Point cloud stitching results after rough calibration.}
\label{Fig.rough_calib}
\end{figure}
The points with small curvature are considered to be in the plane, and then the center point $\textbf{X}$ and normal vector $\textbf{N}$ of the current plane are recorded through the plane fitting. Then the point cloud coordinate transformation casts the point cloud on the first frame. The formula for projecting the point cloud in the LiDAR coordinate system of frame $n$ to GNSS/INS coordinate system of frame $1$ is as follow:

\begin{equation}
\label{equ:coordinate transformation}
    x_{1}^I = (T_{1}^W)^{-1}(T_{W}^I)_n\bm{T_{I}^L}x_{n}^L
\end{equation}
where $x_{n}^L$ represents the n-th frame point cloud in the LiDAR coordinate system.
After being converted to the same coordinate system, the next step is the data association. Similarly, the points on the plane are found through point cloud feature extraction, and then the point cloud in frame $n$ with the plane corresponding to frame 1 and the nearest neighbors of all plane points in frame 1 and frame $n$ are jointly solved for the optimization problem. The extrinsic $\bm{T_{I}^L}$ from LiDAR to GNSS/INS is:

\begin{equation}
\label{equ:resual}
    \bm{T_{I}^L} = argmin{\{\sum_{k=1}^N
    ||\gamma_{plane}^k||^2\}}
\end{equation}

\begin{equation}
\label{equ:r1}
||\gamma_{plane}^k||^2 = \sum_{p=1}^M( \textbf{N}^T(x_p-\textbf{X}))^2
\end{equation}
where 
$||\gamma_{plane}^k||^2$ are the residual errors about 
point to plane. $M$ is the total number of points in the plane. By solving the above least squares problem, we can get an initial solution of the calibration. 

Rough calibration extracts point cloud features and then optimizes the extrinsic parameters according to the features. Fig.~\ref{Fig.rough_calib} shows the surface features extracted by the rough calibration process and the feature point cloud used by the rough calibration to quickly and completely calibrate from the initial state. The calibration speed of this process is very fast. 
We collected multiple calibration sets of data for three different scenes in our experiment, and each data contains 1000+ frame point clouds, the average time of rough calibration is shown in Table \ref{table:rough_time}. The calibration accuracy is shown in Table \ref{table:result}.
In terms of runtime, the calibration can be completed very quickly by extracting features and optimizing, thus saving a lot of runtime, which is conducive to large-scale calibration. In terms of calibration quality, if the requirements are not particularly harsh, the rough calibration result can be used as the final calibration result. 

\subsection{Calibration Refinement}
Through the previous step, we obtained an excellent initial value of extrinsic parameters. To further enhance the effect of mapping, we use octree-based optimization 
to divide the three-dimensional space into a voxel grid. 
We use multi-frame point clouds for splicing and construction, and the point cloud density is relatively dense. In order to improve the calibration speed, we down-sampled the spliced point cloud to a certain extent.
\begin{figure}[ht]
\centering
\includegraphics[scale=0.37]{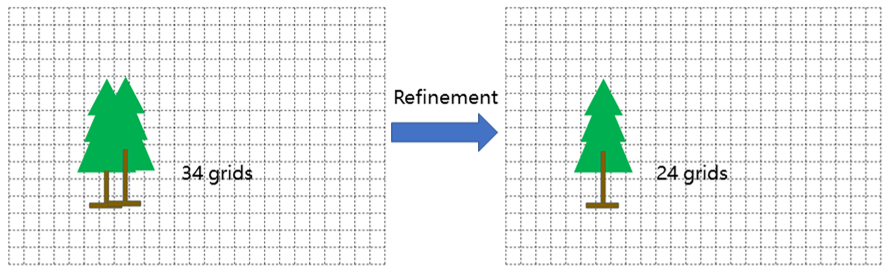}
\caption{Schematic diagram based on space occupancy optimization. For the point cloud bifurcation caused by inaccurate extrinsic parameters, the number of grids occupied in the space is reduced after refinement.}
\label{Fig.octree}
\end{figure}
The goal of our optimization here is the space occupancy rate. The fewer the number of grids occupied by the space, the better the mapping effect and the more accurate the corresponding extrinsic parameters are. First, we use the result of rough calibration to remove the point cloud motion distortion through the uniform speed model.
Then, the initial calibration results are converted to the coordinate system of the first frame through GNSS/INS pose as Eq.\eqref{equ:coordinate transformation}. After the point cloud is transformed into the same coordinate system, the space is divided into a voxel grid. If the calibration result is accurate, the space voxels occupied by all point clouds in the same coordinate system are the smallest, as shown in Fig.~\ref{Fig.octree}. 

\begin{equation}
\label{equ:delT}
    \bm{T_{I}^L}= \arg \min_{T}{\{occupancy(\bm{T_{I}^L},x_p)\}}
\end{equation}
where $x_p$ represents the point cloud after stitching by $\bm{T_{I}^L}$. 
We follow the grid search method described in \cite{levinson2013} to find an optimal calibration parameter so that the point cloud occupies the least number of grids.

\begin{figure}[ht]
\centering
\includegraphics[scale=0.3]{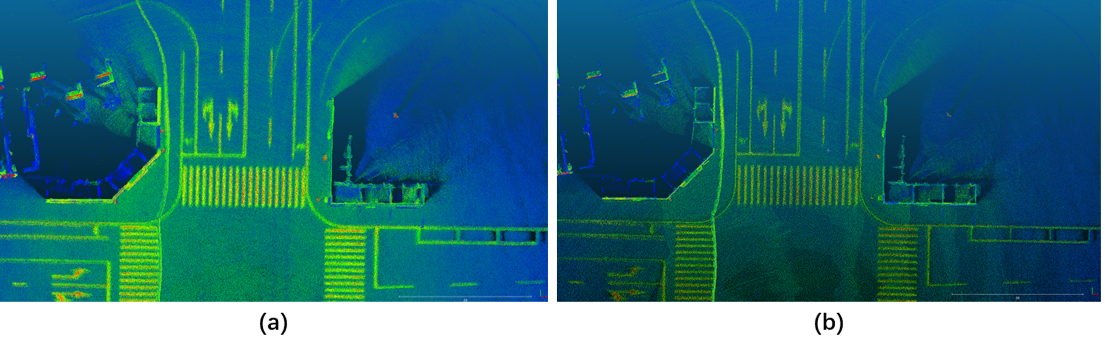}
\caption{The calibration results are further improved by calibration refinement. (a) is the result of rough calibration, (b) is the result of refinement, it can be seen that the wall and lane line become thinner after refinement.}
\label{Fig.refine_calib}
\end{figure}
At the end of the calibration refinement, we get a more accurate calibration result with a better mapping effect.
The octree-based calibration refinement needs to block and traverse the space, so space and time costs are very high. Thanks to the excellent performance of rough calibration, we only optimize within a small extrinsic parameter range in the refine calibration step, so the overall consumption is very low, and the calibration is very fast. 
Table \ref{table:rough_time} shows the average time of refinement calibration.
The result of the calibration refinement is visually similar to the rough calibration in Fig.~\ref{Fig.refine_calib}, the performance at this stage is evaluated by quantitative evaluation in Table \ref{table:result}.

\begin{table}[htbp]
\caption{Calibration Running Time of Three different Scenes}
\centering
\renewcommand{\arraystretch}{1.7}
\begin{tabular}{|c|c|c|c|c}
\hline
 & Scene1 & Scene2 & Scene3 \\
\hline
Our Coarse Calibration & $15.85$ s & $15.80$ s & $13.95$ s \\
\hline
Our Refinement Calibration & $8.64$ s & $8.24$ s & $6.75$ s \\
\hline
Our Z-axis Correction & $0.26$ s & $0.22$ s & $0.17$ s \\
\hline
Method \cite{lidaralign2019} & $179.18$ s & $190.96$ s & $199.79$ s \\
\hline
\end{tabular}  
\label{table:rough_time} 
\end{table}

\subsection{Z-axis Correction}
Because the ground is flat enough for most calibration scenarios, and the excitation in the Z-axis direction is not sufficient in this case, the extrinsic parameter calibration effect of LiDAR and GNSS/INS in the z-axis direction will be terrible. To solve this problem, we propose to use fiducial points to optimize the calibration of the z-axis.
The fiducial points are the three-dimensional point of the accurate world coordinates measured by the positioning device. The number of fiducial points is three or more. It is worth mentioning that each calibration venue only needs to measure the fiducial points once in advance, and any subsequent calibration of vehicles can be performed at this calibration venue without re-measurement of the fiducial points.

\begin{figure}[ht]
\centering
\includegraphics[scale=0.35]{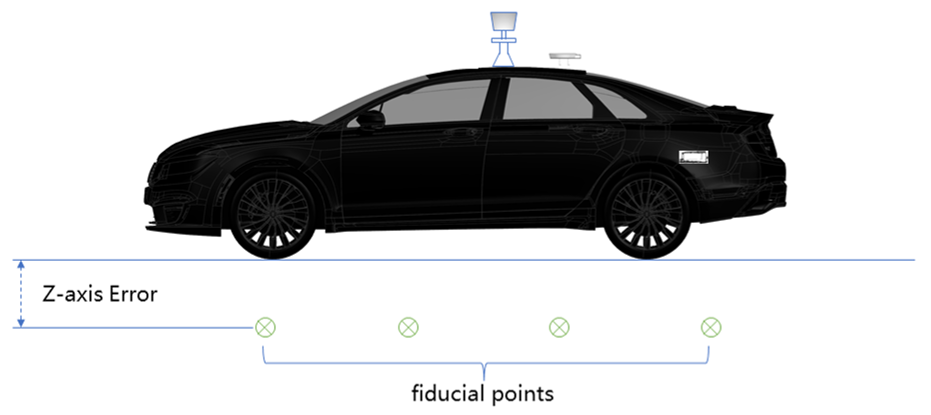}
\caption{The calibration error of the z-axis is further corrected by the fiducial points set in advance on the ground. The reference points in world coordinates are obtained by the 3D coordinate measuring device.}
\label{Fig.z-axis}
\end{figure}

\begin{figure}[ht]
\centering
\includegraphics[scale=0.4]{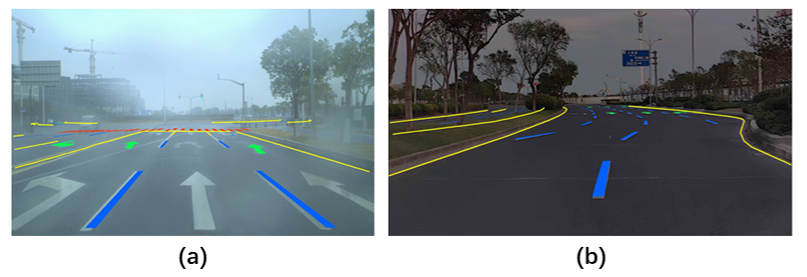}
\caption{(a) is the projection of the map on the image without z-axis correction, and the lane line projection of the map and the lane lines in the image are not aligned. (b) is the projection of the map on the image with z-axis correction.}
\label{Fig.z-axis_correction}
\end{figure}

Finally, we correct the situation on the z-axis. We take $K$ fiducial points of the whole map and project the map to the global coordinate system to build a local map. Then the nearest neighbor of each fiducial point is found on the local map for least square optimization to obtain the final corrected offset on the Z axis. Moreover, this process takes almost no time shown in Table \ref{table:rough_time}. 
The optimization equation is as follows:
 \vspace{-2mm}
\begin{equation}
\label{equ:FIXZ}
    Z_{fix} = \arg \min_{Z} \sum_{i=1}^K{\{||X_{mi}-X_{bi}||^2\}} 
\vspace{-2mm}
\end{equation}
Where $X_{bi}$ represents the i-th fiducial point and $X_{mi}$ represents the nearest neighbor point of the i-th fiducial point in the local map. 
Fig.~\ref{Fig.z-axis} shows the height error between the ground point on the local map and the fiducial point in the actual global coordinate system due to the z-axis error.
Fig.~\ref{Fig.z-axis_correction} shows the projection results of the map with and without Z-axis correction onto the image. If the Z-axis is corrected, it can be perfectly matched. Otherwise, some errors in the z-axis will cause misaligning in height. The performance at this stage is evaluated by quantitative evaluation in Table \ref{table:result}.

\section{EXPERIMENTS}
To evaluate the performance of our method, experiments are conducted on three sets of data. As shown in Fig.~\ref{Fig.calibration_compare}, we selected three calibration scenarios on the road and recorded ten group calibration data for each scenario. The data collection method is shown in  Fig.~\ref{Fig.data_collect} and keeps the vehicle speed between 10 km/h and 20 km/h.
Through qualitative and quantitative evaluation, the results show 
that the proposed method has high accuracy and robustness.


\subsection{Experiment Settings}
We conducted experiments on real driverless platforms, Fig.~\ref{Fig.setup} shows our realistic experiment setup. Top is Hesai Pandar64 LiDAR. GNSS/INS deveice is Novatel PwrPak7 placed in the trunk with two antennas on the roof. 
To evaluate the performance of our method with respect to the reference calibration, we separately measure the error for translation and rotation. 
We also calculate the mean absolute error (MAE) for the three components of translation, namely $\Delta t_x$, $\Delta t_y$, and $\Delta t_z$, as well as the MAE for the three Euler angles $\Delta$\textit{roll}, $\Delta$\textit{pitch}, and $\Delta$\textit{yaw}, which follow the $ZYX$ representation. Our method is implemented in C++ on a desktop computer with an Intel Core i7-8700 CPU and a Nvidia 1660 GPU. 

\subsection{Qualitative Results}
\begin{figure}[ht]
\centering
\includegraphics[scale=0.05]{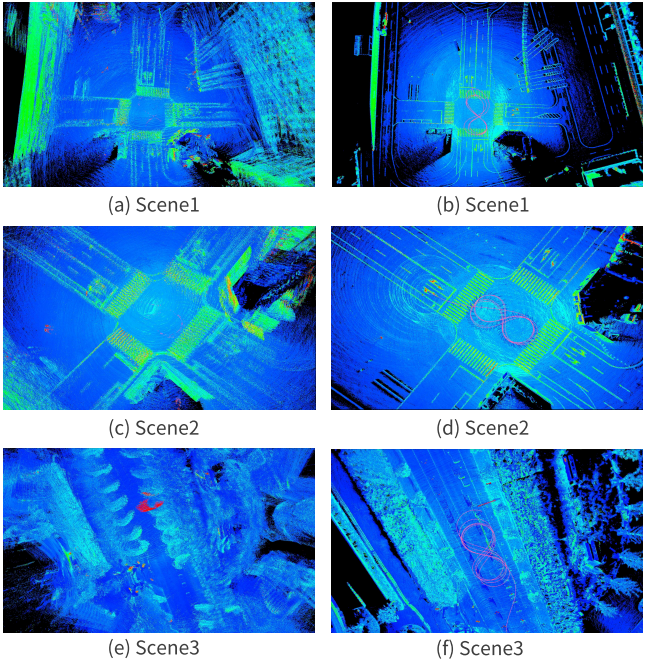}
\caption{The left column in the figure is the point cloud stitching result under the initial inaccurate extrinsic parameters, and the right column is the point cloud stitching result after calibration.}
\label{Fig.calibration_compare}
\end{figure}

\begin{figure}[ht]
\centering
\includegraphics[scale=0.15]{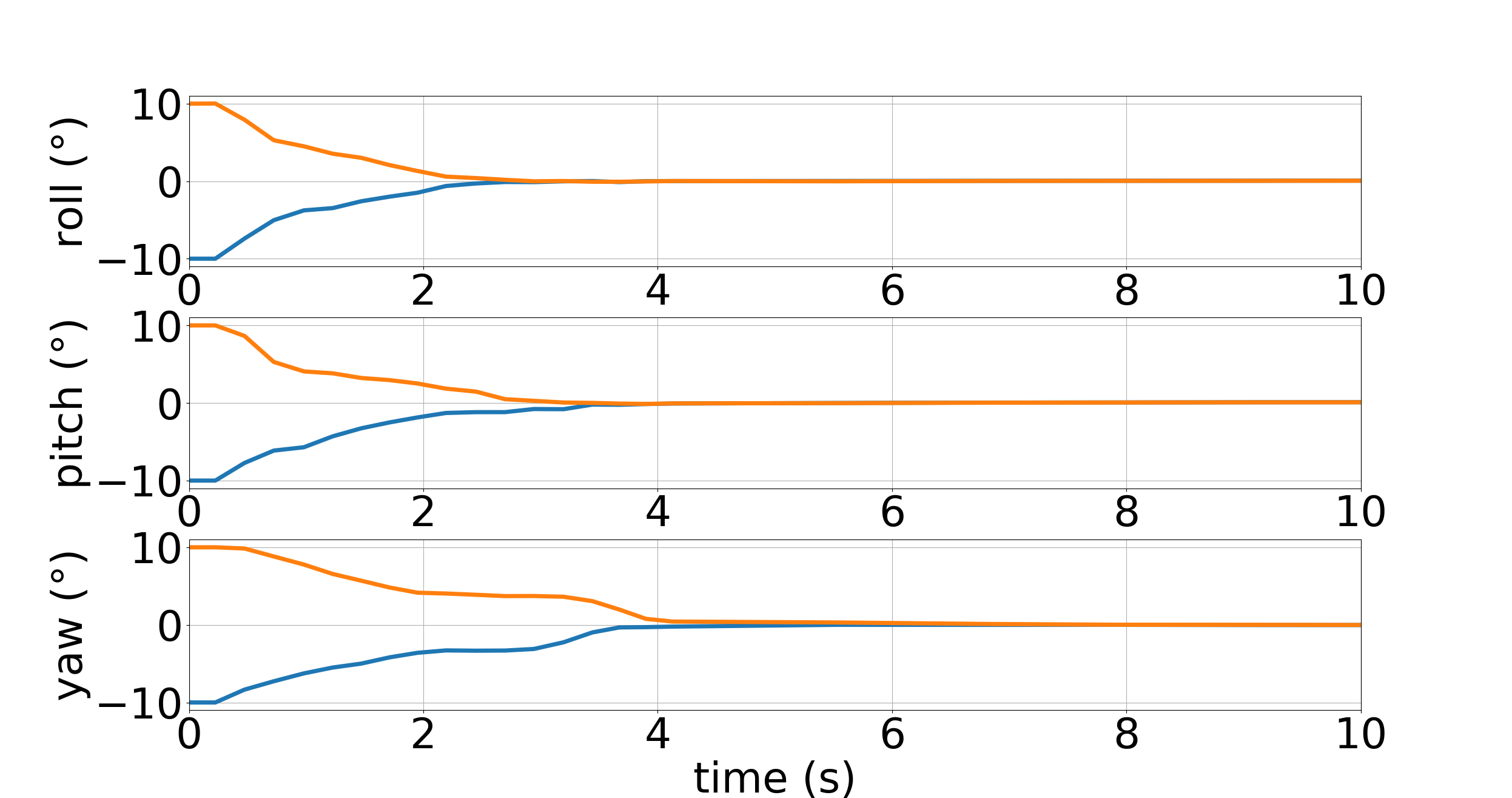}
\caption{In the rough calibration process, the three angles converge to a fixed value with time, and the vertical axis is the error between the current optimization result and the ground truth.}
\label{Fig.convergence_r}
\end{figure}

\begin{figure}[ht]
\centering
\includegraphics[scale=0.15]{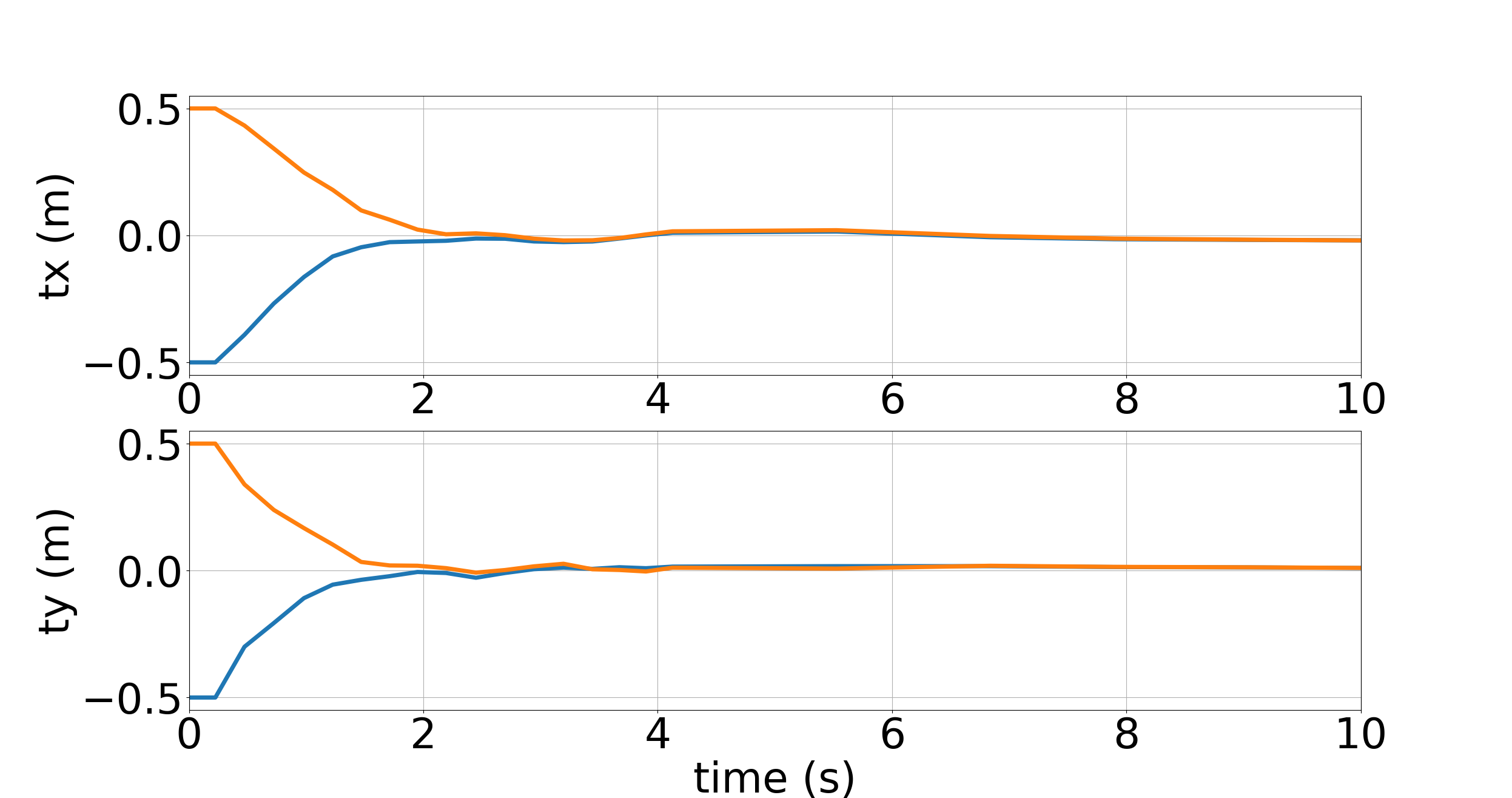}
\caption{In the rough calibration process, the two translation converge to a fixed value with time, and the vertical axis is the error between the current optimization result and the ground truth.}
\label{Fig.convergence_t}
\end{figure}

In order to better visualize the performance of our method, we spliced the point clouds according to the calibration results. Fig.~\ref{Fig.calibration_compare} shows the results before and after calibration. Because the rough calibration results are already excellent, the final and rough calibration results are very similar and hard to see the difference in visualization. In addition, to evaluate the convergence properties of the rough calibration, we gave the initial values of the extrinsic parameters randomly generated in angle and translation from 20° and 0.5m compared to the ground truth.
The whole procedure only takes around $30\text{ s}$ to run in our experiment, and If only the rough calibration process is required, it only takes less than $10\text{ s}$ to converge. 
It is necessary to mention that, to perform rough calibration more robustly, the number of rounds we actually perform calibration is many more than shown in Fig.~\ref{Fig.convergence_r} and Fig.~\ref{Fig.convergence_t}, so the final runtime is about 15s, shown in Table \ref{table:rough_time}. 
Fig.~\ref{Fig.z-axis_correction} shows a comparison of visualizations with and without Z-axis correction, and Z-axis correction reduces the height error with the High-precision map. These experimental results fully demonstrate the robustness and adaptability of our algorithm.

\begin{figure}[ht]
\centering
\includegraphics[scale=0.053]{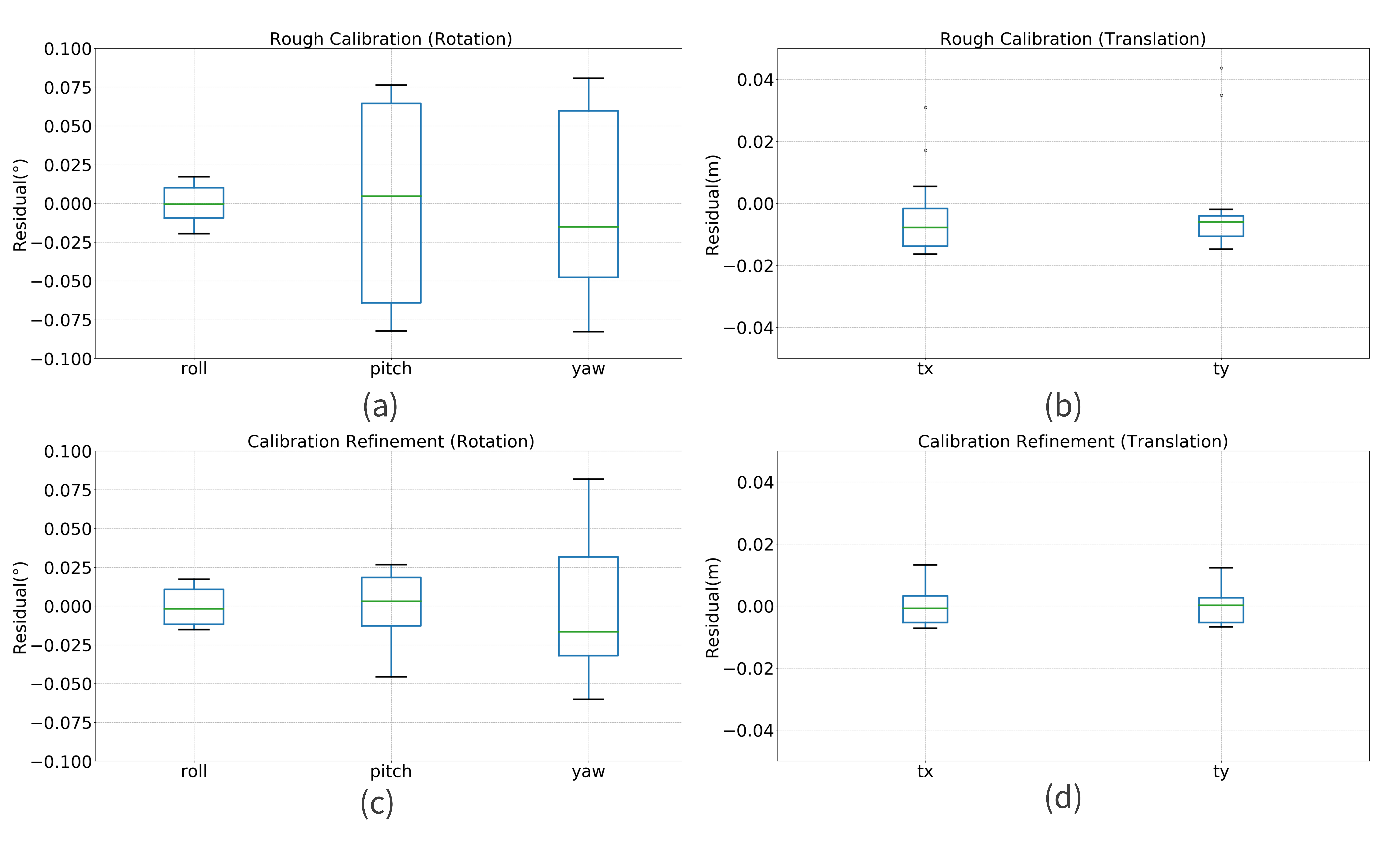}
\caption{(a) and (b) is the consistency evaluation for the rotation and translation of rough calibration. (c) and (d) is the consistency evaluation for the rotation and translation of calibration refinement.}
\label{Fig.consistency_evaluation}
\end{figure}
\subsection{Quantitative Results}

To quantitatively evaluate the error of our calibration method, we obtained a relatively accurate calibration ground truth through the manual tuning tools in \cite{yan2022opencalib}. 
The calibration parameter was verified by various methods, compared with the measured values, and can be taken as the ground truth. 
The MAE is shown in Table \ref{table:result} and our calibration has a higher accuracy.
We also made the distribution statistics on the calibration results of thirty calibration data and obtained the residual vector whose statistical information (e.g., mean, variance) reveals the rough and refine optimization quantitatively. The results are shown in Fig.~\ref{Fig.consistency_evaluation}. It can be seen from the figure that the fluctuation of the calibration extrinsic is smaller for rough and fine calibration, and calibration consistency is also better. Meanwhile, the calibration refinement further improves the calibration accuracy, especially rotation. On the quantitative evaluation of the z-axis, the average error of the z-axis after z-axis correction is kept within $0.015\text{ m}$.

\begin{table}[tbp]
\vspace{-3mm}
\centering
\caption{MAE for Translation and Rotation of Our Method} 
\vspace{-3mm}
 \setlength{\tabcolsep}{2pt}
\renewcommand{\arraystretch}{1.2}
\begin{tabular}{|c|c|c|c|c|c|c|}
\hline
 Calib   & $\Delta t_x$(m) & $\Delta t_y$(m) & $\Delta t_z$(m) & $\Delta$\textit{roll}(\degree) &  $\Delta$\textit{pitch}(\degree) &  $\Delta$\textit{yaw}(\degree) \\ \hline
 Rough  & $0.01705$ & $0.01541$ & $0.35101$ & $0.03742$ & $0.10124$ & $0.13361$     \\ \hline
Refined & $0.00902$ & $0.00716$ & $0.23717$ & $0.01561$ & $0.04878$ & $0.06104$     \\ \hline
Z-axis Correction & $0.00726$ & $0.00793$ & $0.01169$ & $0.00311$ & $0.03426$ & $0.07468$     \\ \hline
Method \cite{lidaralign2019}& $0.21749$ & $0.20326$ & $0.32463$ & $1.49031$ & $1.45612$ & $0.11949$     \\ \hline
\end{tabular}
\label{table:result}
\end{table}

\subsection{Comparison Experiments}
We mainly compared it with \cite{lidaralign2019}, an open-source extrinsic calibration method between a 3D lidar and a 6-dof pose sensor. Table \ref{table:rough_time} and Table \ref{table:result} 
 show the calibration runtime and accuracy comparison using the same calibration initial value on thirty calibration data. 
 At the same time, we found \cite{lidaralign2019} has lower calibration accuracy when the initial value error of the extrinsic parameter is large
and this method does not correct the z-axis calibration accuracy.
In contrast, our method still maintains high accuracy even with poor initial values. Comparative experiments show that our method has advantages in both runtime and calibration accuracy.
\section{CONCLUSIONS}
This paper proposes a three-stage LiDAR to GNSS/INS extrinsic calibration method that maintains high performance in both runtime and accuracy. Our method is specifically designed for autonomous driving and can be applied to the rapid and large-scale calibration of autonomous vehicles. 
This method is optimized by multi-plane features, and the performance may be reduced in the case of relatively few plane features scene. In the future, we will improve our method to adapt to extreme environments with fewer features.

\bibliographystyle{IEEEtran}
\bibliography{egbib}

\end{document}